%% file: main.tex
\documentclass[10pt, a4paper]{article}
\usepackage{lrec}
\usepackage{times}
\usepackage{latexsym}
\usepackage{graphicx}
\usepackage[utf8]{inputenc}
\usepackage{makecell}
\usepackage{microtype}
\usepackage{tabulary}
\usepackage[russian, english]{babel}
\usepackage{color}
\usepackage{url}

\newcommand{\lipart}{\begin{otherlanguage*}{russian}\textit{ли}\end{otherlanguage*}}
\newcommand{\dapart}{\begin{otherlanguage*}{russian}\textit{да}\end{otherlanguage*}}

    \title{SberQuAD -- Russian Reading Comprehension Dataset: \\ Description and Analysis}

\name{Pavel Efimov\textsuperscript{1}\sthanks{~~Work done as an intern at JetBrains Research.}~, Andrey Chertok\textsuperscript{2}, Leonid Boytsov, Pavel Braslavski\textsuperscript{3,4}} 

\address{\textsuperscript{1}Saint Petersburg State University, Saint Petersburg, Russia \\
\textsuperscript{2}Sberbank, Moscow, Russia \\
\textsuperscript{3}Ural Federal University, Yekaterinburg, Russia\\ \textsuperscript{4}JetBrains Research, Saint Petersburg, Russia\\
         pavel.vl.efimov@gmail.com, achertok@sberbank.ru, leo@boytsov.info, pbras@yandex.ru\\
         \\}

\abstract{SberQuAD---a large scale analog of Stanford SQuAD
in the Russian language---is a valuable resource that has not been properly presented to the
scientific community.
We fill this gap by providing
a description, a thorough analysis,
and baseline experimental results.
\\ \newline  }

\begin{document}

\maketitleabstract 

\section{Introduction}
On September 14, 2017 a data science department of Sberbank\footnote{\url{https://www.sberbank.com/about}}---the largest financial institution in Russia---announced a question answering (QA) challenge with substantial monetary prizes.
For this competition Sberbank provided a new large Russian QA dataset containing about 50K training examples, 15K development, and 25K testing examples  (see \S~\ref{SectionDataDesc} for a detailed description).
It was created similarly to the Stanford Question Answering Dataset (SQuAD)~\cite{squad}, which is reflected in its name \textit{SberQuAD} (Sberbank Question Answering Dataset). The competitions had two tasks: retrieval of answer-bearing paragraphs and a reading comprehension (RC) task, which is the focus of this study.

Despite high participation---during a 1.5-month competition 120 participants made 1,348 submissions---the dataset and the contest were neither properly documented nor presented to the scientific community: 
We were able to find only two studies using SberQuAD \cite{kuratov2019adaptation,soboleva2019three}.
Given the importance of the RC task, the scarcity of non-English resources, and the amount of effort went into creation of SberQuAD, it is important to fill the gap.
We in turn provide a historical overview, a description, an in-depth analysis, and baseline experimental results for SberQuAD (using methods previously applied to SQuAD).
We believe this is an important contribution to research in multilingual QA.

\section{Related Work}

QA tasks for unstructured data are typically divided into open-domain  \cite{DBLP:journals/ftir/Prager06} and story comprehension  tasks \cite{DBLP:journals/nle/HirschmanG01}.
In the open-domain setting, to answer a question the system first needs
to guess which documents may contain answers. 
The modern history of open-domain QA starts from TREC challenges organized
by NIST in 2000s \cite{trecQA} and extended by CLEF to a multilingual setting~\cite{clefQA}. 
Notably, in 2011 the open-domain system IBM Watson outstripped two human champions
in the QA contest Jeopardy! \cite{DBLP:journals/aim/FerrucciBCFGKLMNPSW10}.

The story comprehension---commonly referred to as reading comprehension (RC)---is
a more restricted task, where the system needs to answer questions for a given document. 
This task has recently become quite popular with  the introduction 
of a large scale RC dataset named SQuAD \cite{squad}, which was created by crowd workers.
The dataset contains more than 100K questions posed to paragraphs
from popular Wikipedia articles. An answer to each question should
be a valid and relevant paragraph phrase, 
i.e., a 
contiguous sequence of paragraph words including but not limited
to named entities and noun phrases.
The second version of SQuAD (SQuAD 2.0) contains a number of 
unanswerable questions \cite{squad20}.
This makes the task more difficult as the system needs to figure 
out when an answer does not exist.

\input{quest_sample.tex}

Wide adoption of SQuAD led to emergence of many datasets. 
TriviaQA~\cite{TriviaQA} consists of 96K trivia game questions and answers found online
accompanied by answer-bearing documents.
Natural Questions dataset~\cite{naturalquestions} is approximately three times larger than SQuAD.
In that, unlike SQuAD,
questions  are sampled from Google search log
rather than generated by crowd workers.
MS MARCO~\cite{msmarco} contains 1M questions from 
a Bing search log along with free-form answers.
For both MS MARCO and Natural Questions answers
are produced by in-house annotators.
QuAC~\cite{QuAC} and CoQA~\cite{CoQA} contain questions and answers in information-seeking dialogues.
For a more detailed discussion we address the
reader to a recent survey~\cite{DBLP:journals/corr/abs-1907-01686}.

Majority of RC dataset are in English. Few exceptions are Chinese datasets  WebQA~\cite{WebQA} and DuReader~\cite{dureader}, as well as Bulgarian~\cite{hardalov2019englishonly} and  Tibetian~\cite{TibetanQA} ones. Recently,~\newcite{artetxe2019cross} experimented with cross-language transfer learning and prepared XQuAD dataset containing 240 paragraphs and 1,190 question-answer pairs from SQuAD v1.1 translated into 10 languages.  

A number of studies scrutinize existing datasets to evaluate the difficulty of the task
as well as robustness of the models. \newcite{chen2016thorough} sample 100 passage-question-answer triples from CNN/Daily Mail dataset~\cite{CNN_dailymail}, classify them manually according to difficulty into several categories, and compare performance of several models for different levels of question complexity. \newcite{jia2017adversarial} generate semi-automatically distracting sentences into paragraphs to investigate the robustness of neural reading comprehension models.
\newcite{talmor2019multiqa} study how well neural network models transfer among different RC datasets. 
\newcite{wadhwa2018comparative} perform quantitative and qualitative analysis of four neural models on SQuAD. 
This work is close to ours, though our focus is on exploring the SberQuAD dataset, rather than models.

\newcite{evaluation-metrics} 
evaluate dataset difficulty from a human perspective. 
They compute readability scores and the number of skills,
such as reasoning and co-reference capability, required to answer questions for six datasets.
They show that SQuAD is an easy-to-answer but 
hard-to-read dataset, which requires few skills to answer questions.
\newcite{systematic-error-analysis}
argue that human explanations are not reliable
and instead estimate question complexity based on 
performance of several models:  
The fewer models can answer a question, 
the higher is its complexity. 
They find that complexity correlates 
with several features including the presence
of named entities and the density of question words
in the answer, but not with readability scores.
In particular, if the answer is an named entity
the question is easy in 72\% of the cases
as opposed to 44\% of the cases when the answer is not an entity.

\section{Dataset}\label{SectionDataDesc}

In this paper, we focus on task B of the Sberbank 2017 competition and the corresponding RC dataset, namely, SberQuAD. 
Details of the competition can be found on the competition website\footnote{\url{https://github.com/sberbank-ai/data-science-journey-2017}}.
The original dataset comes in CSV format:
A variant in SQuAD JSON format can be downloaded from the \texttt{DeepPavlov} QA project page.\footnote{\url{http://docs.deeppavlov.ai/en/master/features/models/squad.html} DeepPavlov is an open NLP framework maintained by the Neural Networks and Deep Learning Lab of Moscow Institute of Physics and Technology (MIPT), see~\cite{deeppavlov}.} 
A lively discussion of approaches and their effectiveness can be found in the OpenDataSciene community slack\footnote{\url{https://opendatascience.slack.com/}, hashtag \url{#sberbank_contest}, September--November 2017.}.
The list of top-10 best performing teams can be found in the video of the celebration ceremony.\footnote{\url{https://youtu.be/J5HOjC4Xn_Y?t=29830}}

As we learned from a private communication with the dataset developers, 
they generally followed the procedure described in the SQuAD paper \cite{squad}.
First they selected Wikipedia pages,
split them into paragraphs,
and presented paragraphs to crowd workers.
For each paragraph,
a crowd worker had to come up with questions that can be 
answered using solely the content of the paragraph.
In that, an answer must have been a paragraph span, i.e.,
a contiguous sequence of paragraph words.
The tasks were posted on Toloka\footnote{\url{https://toloka.yandex.com}} crowdsourcing platform. 
SberQuAD contains 50,364 paragraph--question--answer triples in the training set which are publicly available. 
A development set, which was available only during the competition, contains about 15K triples and the hold-out test containing 25K triples is kept private. 
There are two differences between SQuAD and SberQuAD formats.
First, SberQuAD does not tell us which Wikipedia pages a paragraph belongs to.
Second, each answer is represented by a string without respective starting position in the paragraph.

\input{datastat_table.tex}

\paragraph{Examples and basic statistics.}
Figure~\ref{fig:examples} shows a sample SberQuAD paragraph with three questions:
Gold-truth answers are underlined in text.
Generally, the format of the question and the answers mimics that of SQuAD v1.1.
Note, however, the following peculiarities:
Question \textit{Q30330} contains a spelling error; 
Question \textit{Q28900} references prior question \textit{Q11870}
and cannot, thus, be answered on its own (likely both questions were created by the same crowd worker).

\begin{figure}[t]
    \center
    \includegraphics[width=.4\textwidth]{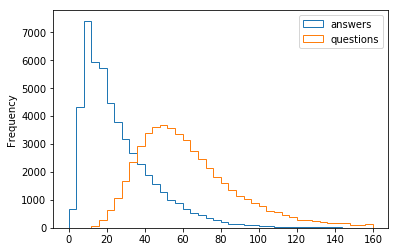}
    \caption{Question/answer length histograms (in chars)}
    \label{fig:QA_length_distr}
\end{figure}

Basic dataset statistics is summarized in Table~\ref{tab:data_stats}: 
SberQuAD has about twice as fewer questions compared to SQuAD
whereas the average lengths of  paragraphs, questions, and answers are very similar for both datasets.
Distribution of question/answer length is presented in Figure~\ref{fig:QA_length_distr}.
There are 275 questions (0.55\%) having at least 200 characters
and 374 answers (0.74\%) that are longer than 100 characters. 
Questions are substantially longer than answers.
Anecdotally, very long answers and very short questions are frequently errors. 
For example, for question \textit{Q61603} the answer field contains a copy of the whole paragraph,
while question \textit{Q76754} consists of a single word `thermodynamics'.

Because the original SberQuAD development set is not available,
the original training set of SberQuAD was partitioned into a (new) training (45,328) and testing (5,036) sets by the \texttt{DeepPavlov} team.
This is the partition that we use in our experiments: We train models on the training set and evaluate them
on the testing set.

\paragraph{Analysis of questions.}

Most questions in the dataset start with either a question word or preposition:
ten most common starting words are \foreignlanguage{russian}{что} (what), \foreignlanguage{russian}{в} (in), \foreignlanguage{russian}{как} (how), \foreignlanguage{russian}{кто} (who), \foreignlanguage{russian}{какие} (what adj), \foreignlanguage{russian}{когда} (when), \foreignlanguage{russian}{какой} (what adj), \foreignlanguage{russian}{где} (where), \foreignlanguage{russian}{сколько} (how many), \foreignlanguage{russian}{на} (on).

These starting words correspond to 62.4\% of all questions.
In about 4\% of the cases, an interrogative word is not among the first three words of the question, though.
Manual inspection showed that in most cases these entries are declarative statements, 
sometimes followed by a question mark, e.g. \textit{Q15968 `famous Belgian poets?'}, or ungrammatical questions. 

To get a better understanding of question types, 
we inspected questions' most common lemmatized starting bigrams (Table~\ref{tab:bigrams_answer_types}) and trigrams (Table  \ref{tab:trigrams_answer_types}). 
In Russian, an interrogative word is often preceded by a preposition, which results in a high variability of starting n-grams: 
As one can see from the Table~\ref{tab:bigrams_answer_types}, ten most frequent bigrams account only for about 19\% of all questions. 
Judging by bigram statistics, 
definition  (\textit{what do you call/what is X}) and time-related questions (\textit{when}) are among most popular ones. 
Trigram statistics (Table~\ref{tab:trigrams_answer_types}) 
permits a more precise inference about most common question types: 
They include variations of time-related questions (\textit{in which year/century/period}), 
 location questions (\textit{in which city/country}), as well as causality questions (\textit{what does X lead to/what does X depend on}).

\paragraph{Analysis of answers.}

While manually examining the dataset, we encountered misspelled questions. 
To estimate the proportion of questions with misspellings, 
we verified all questions using Yandex spellchecking API\footnote{\url{https://yandex.ru/dev/speller/} (in Russian)}. 
The automatic speller identified 2,646 and 287 misspelled questions in training and testing sets, respectively. 
According to a manual assessment of 200 randomly selected questions, the spellchecker has precision 0.62.\footnote{This score is significantly lower than scores  obtained for Yandex spellchecker by \texttt{DeepPavlov} ($P=83.09; R=59.86$), see \url{http://docs.deeppavlov.ai/en/master/features/models/spelling_correction.html}} 
Manual inspection suggests that most false positives are due to either spelling/inflectional variants
or rare words being replaced with more frequent ones (apparently based on language model scores).

We also found 385 and 51 questions in training and testing sets, respectively, containing Russian interrogative particle \lipart\        (\textit{whether/if}).
This form implies a yes/no question,
which is generally not possible to answer in the RC setting by selecting a valid and relevant paragraph phrase. 
For this reason, most answers for these yes-no questions are fragments supporting or refuting the question statement. 
In addition, we found 15 answers in the training set, where the correct answer `yes' (Russian \dapart) 
can be found as a paragraph word substring, but not as a valid/relevant phrase.

\begin{table}[tb]
    \centering
    \begin{tabular}{lrrr}
        \textbf{NE} &  \textbf{Manual} & \textbf{DPNER (P/R)}  & \textbf{Exact}\\
         \hline
        Date         & 9.9\%  & 12.6\% (0.83/0.96) & 2.31\%\\
        Number      & 11.0\%  & 9.9\% (0.85/0.90) & 3.37\%\\
        Person       & 8.8\%  & 8.2\% (0.89/0.89) & 3.85\%\\
        Location     & 5.4\%  & 7.6\% (0.64/0.87) & 1.45\%\\
        Organization & 4.0\%  & 3.6\% (0.70/0.70) & 1.46\%\\
        Other NEs     & 5.2\%  & 2.5\% (0.71/0.44) & 0.97\%\\
    \end{tabular}
    \caption{Named entities in a manually annotated sample of 1,000 answers (\textbf{Manual}); answers containing automatically detected NEs (\textbf{DPNER}); detection quality on manually annotated sample (\textbf{P/R}); automatically detected NEs that exactly match answers' boundaries (\textbf{Exact}).}
    \label{tab:NER_quantity_quality}
\end{table}{}

Following~\cite{squad}, we analyzed answers presented in the dataset by their type.
To this end, we employed a NER tool from \texttt{DeepPavlov} library, DPNER hereafter.\footnote{\url{http://docs.deeppavlov.ai/en/master/features/models/ner.html}} DPNER is a multilingual BERT model trained on OntoNotes corpus annotated with 19 entity types and transferred to Russian (for 
a discussion of zero-shot transfer see a paper by \newcite{DBLP:conf/acl/PiresSG19}).
To evaluate DPNER on SberQuAD data, we randomly sampled 1,000 answers and manually 
tagged containing named entities (NE) using
the following tags: DATE, NUMBER, PERSON, LOCATION, ORGANIZATION, and OTHER (artwork, TV show, etc.).
When an answer contained more than one entity,
we highlighted the entity representing a key answer concept (e.g., the head noun phrase).
For example \textit{26-year-old Comtesse Sophie d'Houdetot} (Q56395) was marked as PERSON.

This statistics is summarized in Table ~\ref{tab:NER_quantity_quality}.
The first column shows distribution of NEs in answers according to manual annotation.
The second column shows precision/recall of the NER tool applied to sentences containing answers.
The last column of the Table~\ref{tab:NER_quantity_quality} 
reports a fraction of answers (identified by DPNER)
that are NE (as opposed only a part of the answer being a NE). 
In total, DPNER found NEs in almost 43\% of answers in the dataset.

\begin{table}[t]
    \centering
    \begin{tabular}{l@{\hspace{0em}}r@{\hspace{0.25em}}r@{\hspace{0.25em}}r@{\hspace{0.25em}}r@{\hspace{0.25em}}r@{\hspace{0.25em}}r}
        \textbf{Type} &  \textbf{\% test} & \textbf{R-Net} & \textbf{BiDAF} & \textbf{DocQA} & \textbf{DrQA} & \textbf{BERT}\\
        \hline
        NP              & 24.0 & 77.5 & 70.3 & 78.2 & 73.5 & 84.5 \\
        PP              & 10.5 & 83.1 & 78.6 & 84.9 & 81.4 & 89.1 \\
        VP              &  7.1 & 61.9 & 54.0 & 62.7 & 55.5 & 71.6 \\
        ADJP            &  5.9 & 73.0 & 65.3 & 75.5 & 67.2 & 80.5 \\
        ADVP            &  0.3 & 67.9 & 45.3 & 70.7 & 51.2 & 76.6 \\
        non-R           &  0.3 & 91.7 & 88.2 & 98.2 & 92.9 & 95.1 \\
        None            &  9.1 & 75.7 & 69.0 & 77.1 & 70.1 & 83.0 \\
        \hline
         Test set               & &     77.8 & 72.2 & 79.5 & 75.0 & 84.8\\
    \end{tabular}
    \caption{Distribution of answers by constituent types (NP -- noun phrase, PP -- prepositional phrase, VP -- verb phrase,  ADJP -- adjective phrase, ADVP -- adverb phrase, non-R -- words in non-Russian characters; None -- not recognized).}
    \label{tab:answers_AOT}
\end{table}

Following~\cite{squad}, we complemented our analysis of answers with syntactic parsing.
To this end we applied the rule-based constituency parser AOT \footnote{\url{http://aot.ru}} to answers without detected NE.
When AOT produced multiple parses, 
we picked the parse with the longest span. 
AOT parser supports a long list of phrase types (57 in total),
we grouped them into conventional high-level types, which are shown in Table~\ref{tab:answers_AOT}
\footnote{Table~\ref{tab:answers_AOT} provides data for the testing set, but the distribution for the training set is quite similar.}.  Not surprisingly, noun phrases are most frequent answer types, followed by prepositional phrases. 
Verb phrases represent a non-negligible share of answers (7.1\%), 
which is quite different from a traditional 
QA setting where answers are predominantly noun phrases \cite{DBLP:journals/ftir/Prager06}. 

\begin{figure}[t]
    \center
    \vspace{-1.25em}
    \includegraphics[width=.4\textwidth]{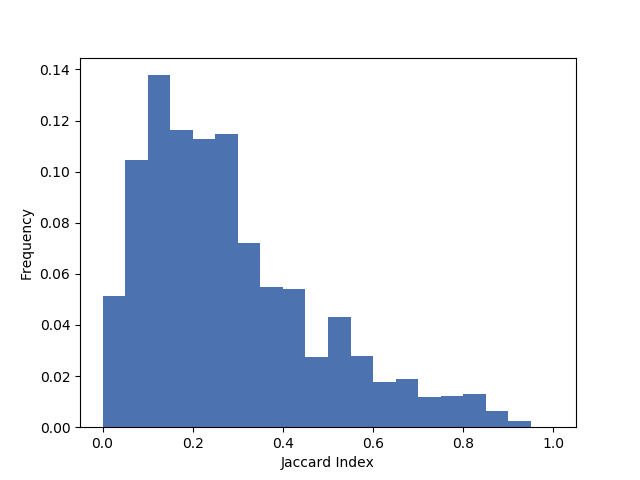}
    \caption{Jaccard similarity distribution between questions and answer containing sentences.}
    \label{fig:jaccard_unigrams}
\end{figure}

\paragraph{Question/paragraph similarity.} We further estimate similarity 
between questions and paragraph sentences containing the answer:
The more similar is the question to its answer's context, the simpler is the task of locating the answer. 
In contrast to SQuAD we refrain from syntactic parsing and rely on simpler approaches. 
First, we compared questions with complete paragraphs.
To this end, we calculated the length of the the longest contiguous matching subsequence (LCMS) 
between a question and a paragraph using the \texttt{difflib} library.\footnote{\url{https://docs.python.org/3/library/difflib.html}} 
The last row in Table~\ref{tab:data_stats} shows that despite similar paragraph and question lengths in both SQuAD and SberQuAD,
the SberQuAD questions are more similar to the paragraph text. 

Second, we estimated similarity between a question and the sentence containing the answer.
To ensure the accuracy of estimation,
we evaluated several available tools for sentence splitting on a random sample of 100 SberQuAD paragraphs,
which were manually split into 590 sentences.
\texttt{DeepPavlov} tokenizer\footnote{\url{https://github.com/deepmipt/ru_sentence_tokenizer}} outperformed other tools in terms of quality ($P/R =0.93/0.94$) and efficiency, so we applied it to the whole train subset. 
Subsequently, we lemmatized the data using \textit{mystem}\footnote{\url{https://yandex.ru/dev/mystem/} (in Russian)} 
and calculated the Jaccard coefficient between 
a question and the sentence containing the answer. 
The distribution of the scores is presented in Figure~\ref{fig:jaccard_unigrams}.
The mean value of the Jaccard coefficient is 0.28 (median is 0.23). 

Our analysis shows that there is a substantial lexical overlap between questions and paragraph sentences containing the answer, 
which may indicate a heavier use of the copy-and-paste approach by crowd workers recruited for SberQuAD creation. \footnote{Note that in the interface for crowdsourcing SQuAD questions, prompts at each screen reminded the workers to formulate questions in their own words; in addition, the copy-paste functionality for the paragraph was purposefully disabled.} 

\begin{table}[t]
    \centering
    \begin{tabular}{l@{\hspace{0em}}r@{\hspace{0.25em}}r@{\hspace{0.25em}}r@{\hspace{0.25em}}r@{\hspace{0.25em}}r@{\hspace{0.25em}}r}
        \textbf{NE} & \textbf{\% test} &  \textbf{R-Net} & \textbf{BiDAF} & \textbf{DocQA} & \textbf{DrQA} & \textbf{BERT}\\
         \hline
        Date         & 12.2\% & 88.0 & 86.6 & 90.0 & 88.9 & 91.3 \\
        Number       &  9.6\% & 73.1 & 69.1 & 75.5 & 72.5 & 80.4 \\
        Person       &  8.8\% & 78.3 & 73.1 & 81.0 & 77.7 & 86.6 \\
        Location     &  7.6\% & 79.8 & 75.7 & 81.1 & 77.8 & 85.8 \\
        Organization &  4.1\% & 79.0 & 77.3 & 82.3 & 78.3 & 88.2 \\
        Other NE     &  2.1\% & 72.7 & 59.4 & 73.6 & 64.7 & 80.9 \\
        \hline
        Any NE       & 42.7\% & 80.3 & 76.4 & 82.6 & 79.7 & 87.0 \\
        \hline
        Test set                & &     77.8 & 72.2 & 79.5 & 75.0 & 84.8\\
    \end{tabular}
    \caption{Model performance on answers containing named entities.}
    \label{tab:NE_answer_types}
\end{table}{}

\pagebreak

\section{Employed Models}
We used the following models:
\begin{itemize}
    \item Two baselines provided by SberQuAD organizers;
    \item Four models, which have strong performance among models not relying on transformers. They were used in a study similar to ours~\cite{wadhwa2018comparative};
    \item BERT model provided by the \texttt{DeepPavlov} library, which employs
          large pre-trained transformers \cite{bert,DBLP:conf/nips/VaswaniSPUJGKP17}.
\end{itemize}

\paragraph{Preprocessing and training.}
We tokenized text using spaCy\footnote{\url{https://github.com/buriy/spacy-ru}}.
To initialize the embedding layer for BiDAF, DocQA, DrQA, and R-Net
we use Russian case-sensitive fastText embeddings trained on Common Crawl and Wikipedia\footnote{\url{https://fasttext.cc/docs/en/crawl-vectors.html}}.
This initialization is used for both questions and paragraphs.
For BiDAF and DocQA about 10\% of answer strings in both training and testing sets require a correction of positions,
which can be nearly always achieved automatically by ignoring punctuation (12 answers required a manual intervention).
Models were trained on GPU nVidia Tesla V100 16Gb.
We used default implementation settings, which are listed in
Table \ref{tab:TrainParam}:

\begin{table}[h]
\centering
\begin{tabular}{l@{\hspace{0.25em}}c@{\hspace{0.25em}}c@{\hspace{0.25em}}c@{\hspace{0.25em}}c@{\hspace{0.25em}}}
 Model  & Optim. & Batch & \# epochs & Init. LR \\\hline
 R-Net  & Adadelta & 32 & ~40 (60K steps) & 0.5 \\
 BiDAF  & Adadelta & 60 & 12              & 0.5 \\
 DocQA  & Adadelta & 45 & 26              & 1 \\
 DrQA   & Adamax   & 128 & 40             & N/A \\\hline
\end{tabular}
\caption{Training parameters. LR stands for learning rate.\label{tab:TrainParam}}
\end{table}

\paragraph{Baselines.} Contest organizers made two baselines\footnote{\url{https://github.com/sberbank-ai/data-science-journey-2017/tree/master/problem_B/}} available.
\textbf{Simple baseline:} The model returns a sentence with the maximum word overlap with the question.
\textbf{ML baseline} generates features for all word spans in the sentence returned by the \textbf{simple baseline}.
The feature set includes TF-IDF scores, span length, distance to the beginning/end of the sentence, as well as POS tags. 
The model uses gradient boosting to predict F1 score. At the testing stage the model selects a candidate span with maximum predicted score.

\paragraph{Gated Self-Matching Networks (R-Net):}
This model, proposed by~\newcite{r-net2017}, is a multi-layer end-to-end neural network that uses a gated attention mechanism to give different levels of importance to different paragraph parts. It also uses self-matching attention for the context to aggregate evidence from the entire paragraph to refine the query-aware context representation. We use a model implementation by HKUST\footnote{\url{https://github.com/HKUST-KnowComp/R-Net}}. To increase efficiency, the implementation adopts scaled multiplicative attention instead of additive attention and uses variational dropout.

\paragraph{Bi-Directional Attention Flow (BiDAF):}
The model proposed by~\newcite{biDAF} takes inputs of different granularity (character, word and phrase) to obtain a query-aware context representation without previous summarization using memory-less context-to-query (C2Q) and query-to-context (Q2C) attention. We use original implementation by AI2\footnote{\url{https://github.com/allenai/bi-att-flow}}.

\begin{table}[t]
    \centering
    \begin{tabular}{l@{\hspace{0em}}r@{\hspace{0.25em}}r@{\hspace{0.25em}}r@{\hspace{0.25em}}r@{\hspace{0.25em}}r@{\hspace{0.25em}}r}
        \textbf{NE} & \textbf{\% test} &  \textbf{R-Net} & \textbf{BiDAF} & \textbf{DocQA} & \textbf{DrQA} & \textbf{BERT}\\
         \hline
        Date         & 2.2\%  & 87.1 & 87.3 & 90.8 & 87.5 & 95.0 \\
        Number       &  3.3\% & 78.2 & 72.4 & 80.1 & 77.7 & 90.2 \\
        Person       &  4.2\% & 83.2 & 74.0 & 85.1 & 82.9 & 91.4 \\
        Location     &  1.7\% & 78.3 & 72.8 & 82.1 & 77.9 & 88.6 \\
        Organization &  1.5\% & 80.7 & 76.5 & 81.6 & 79.2 & 91.8 \\
        Other NE     &  0.9\% & 80.9 & 54.9 & 78.1 & 66.4 & 88.9 \\
        \hline
        Any NE       & 13.8\% & 81.6 & 74.5 & 83.6 & 80.2 & 91.2 \\
        \hline
        Test set    & &     77.8 & 72.2 & 79.5 & 75.0 & 84.8\\
    \end{tabular}
    \caption{Model performance on answers matching NER tags.}
    \label{tab:NE_whole_answer_types}
\end{table}{}

\paragraph{Multi-Paragraph Reading Comprehension (DocQA):}
This model, proposed by \newcite{docqa}, aims to answer questions based on entire documents (multiple paragraphs). If considering the given paragraph as the document, it also shows good results on SQuAD. It uses the bi-directional attention mechanism from the BiDAF and a layer of residual self-attention. We also use original implementation by  AI2\footnote{\url{https://github.com/allenai/document-qa}}.

\paragraph{Document Reader (DrQA):}
This model proposed by ~\newcite{drqa} is part of the system for answering open-domain factoid questions using Wikipedia. The Document Reader component performs well on SQuAD (skipping the document retrieval stage). The model has paragraph and question encoding layers with RNNs and an output layer. The paragraph encoding passes as input to RNN a sequence of feature vectors derived from tokens: word embedding, exact match with question word, POS/NER/TF and aligned question embedding. The implementation is developed by Facebook Research\footnote{\url{https://github.com/facebookresearch/DrQA}}.

\paragraph{Bidirectional Encoder Representations from Transformers (BERT):}
We use a BERT-based QA model by \texttt{DeepPavlov}\footnote{\url{http://docs.deeppavlov.ai/en/master/features/models/squad.html}}. Pre-trained BERT models achieved superior performance is a variety of downstream NLP tasks, including RC~\cite{bert}.
The Russian QA model is obtained by a transfer from the multilingual BERT (mBERT) 
with subsequent fine-tuning on the Russian Wikipedia and SberQuAD \cite{kuratov2019adaptation}.

\paragraph{Evaluation.} Similar to SQuAD, SberQuAD evaluation employs two metrics to assess model performance -- 1) the percentage of system's answers that exactly match (EM) any of the gold standard answers and 2) the maximum overlap between the system response and ground truth answer at the token level expressed via F1 (averaged over all questions). 
Both metrics ignore punctuation and capitalization. 

\begin{table}[t]
\begin{center}
\begin{tabular}{l|rr|rr}
\textbf{Model} & \multicolumn{2}{c|}{\textbf{SberQuAD}} & \multicolumn{2}{c}{\textbf{SQuAD}}\\ 
& EM & F1 & EM & F1 \\
\hline
simple baseline  & 0.3 & 25.0 & -- & -- \\
ML baseline &  3.7 & 31.5 & -- & -- \\
BiDAF &  51.7 & 72.2 & 68.0 & 77.3 \\
DrQA  & 54.9 & 75.0 & 70.0 & 79.0 \\
R-Net  & 58.6 & 77.8 & 71.3 & 79.7 \\
DocQA &  59.6 & 79.5 & 72.1 & 81.1 \\
BERT & 66.6 & 84.8 & 85.1 & 91.8\\
\end{tabular}
\end{center}
\caption{\label{tab:main_results} Model performance on SQuAD and SberQuAD; 
SQuAD part shows single-model scores on test set taken from respective papers.}
\end{table}

\begin{figure}[t]
    \center
    \includegraphics[width=0.4\textwidth]{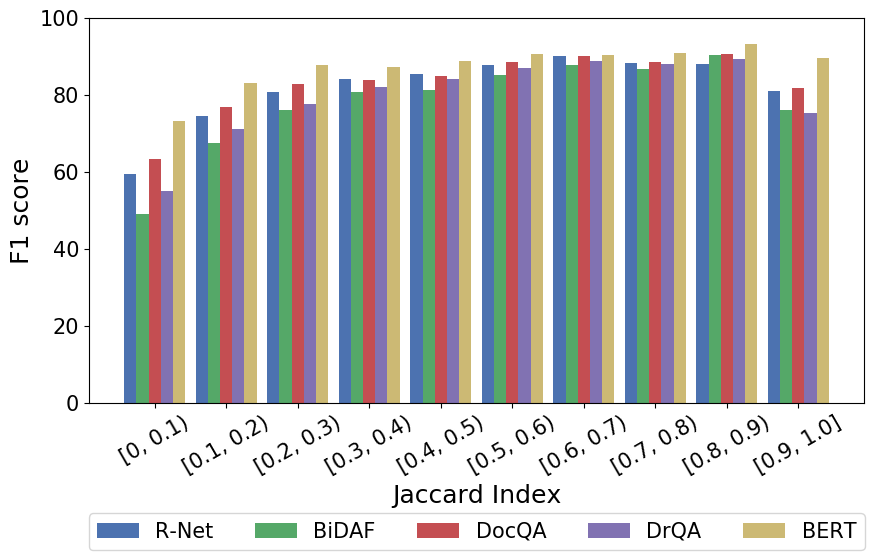}
    \caption{Model performance depending on Jaccard similarity between a question and the sentence containing an answer.}
    \label{fig:jaccard_and_f1}
\end{figure}

\section{Analysis of Model Performance}

Main experimental results are shown in Table~\ref{tab:main_results}. 
It can be seen that all the models perform worse on the Russian dataset SberQuAD than on SQuAD.
In that, there is a bigger difference in exact matching scores compared to F1.
For example, for BERT the F1 score drops from 91.8 to 84.8 whereas the exact match score drops from  85.1 to 66.6.
The relative performance of models is consistent for both datasets, 
although there is a greater variability among four neural ``pre-BERT'' models.
One explanation for lower scores is that SberQuAD has always only one 
correct answer, whereas SQuAD can have multiple answer variants (1.7 on the development set).
Furthermore, SberQuAD contains many fewer answers that are named entities than SQuAD (13.8\% vs. 52.4\%),
which---as we discuss below---maybe another reason for lower scores.
Another plausible reason is a poorer quality of annotations: 
We have found a number of deficiencies including but not limited to misspellings in questions and answers.

Figure~\ref{fig:jaccard_and_f1} shows the relationship between the F1 score and the question-answer similarity
expressed as the Jaccard coefficient.
Note that 64\% of question--sentence pairs fall into first three bins.
As expected, a higher value of the Jaccard coefficient corresponds to higher F1 scores 
(with the exception of 14 questions where Jaccard is above 0.9).\footnote{Among these 14 questions the majority are long sentences from the paragraph with a single word (answer) substituted by a question word; there is an exact copy with just a question mark at the end; one question has the answer erroneously attached after the very question.}
Furthermore, in the case of the high similarity there is only a small difference among model performance.
These observations support the hypothesis that it is easier to answer questions when there is 
 a substantial lexical overlap between a question and a paragraph sentence containing the answer.

\begin{figure}[t]
    \center
    \includegraphics[width=0.4\textwidth]{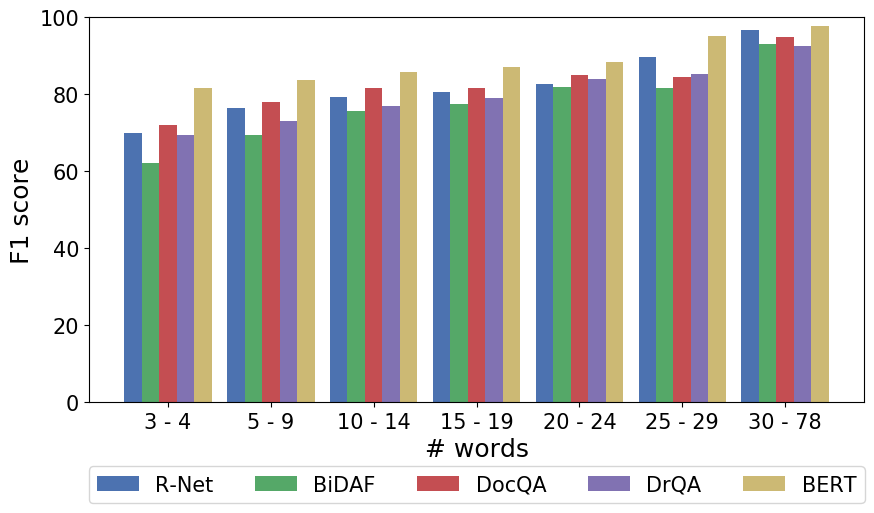}
    \caption{Model performance depending on question length (\# of words).}
    \label{fig:q_length_and_f1}
\end{figure}

Longer questions are easier to answer too: 
According to Figure~\ref{fig:q_length_and_f1},
the F1 score increases nearly monotonically with the question length.
Presumably, longer questions provide more context for identifying correct answers. 
In contrast, dependency on the answer length is not monotonic:
the F1 score first increases and achieves the maximum for 2-4 words.
A one-word ground truth constitutes a harder task: 
missing a single correct word results in a null F1 score, whereas returning a two-word answer containing the single correct word results in only $F1=0.67$.
F1 score also decreases substantially for answers above average length.
It can be explained by the fact that models are trained on the dataset where shorter answers prevail, see Table~\ref{tab:data_stats} and Figure~\ref{fig:QA_length_distr}. Models' average-length answers get low scores in case of longer ground truth. For example, a 4-word answer fully overlapping with a 8-word ground truth answer gets again only $F1=0.67$.

\begin{figure}[t]
    \center
    \includegraphics[width=0.4\textwidth]{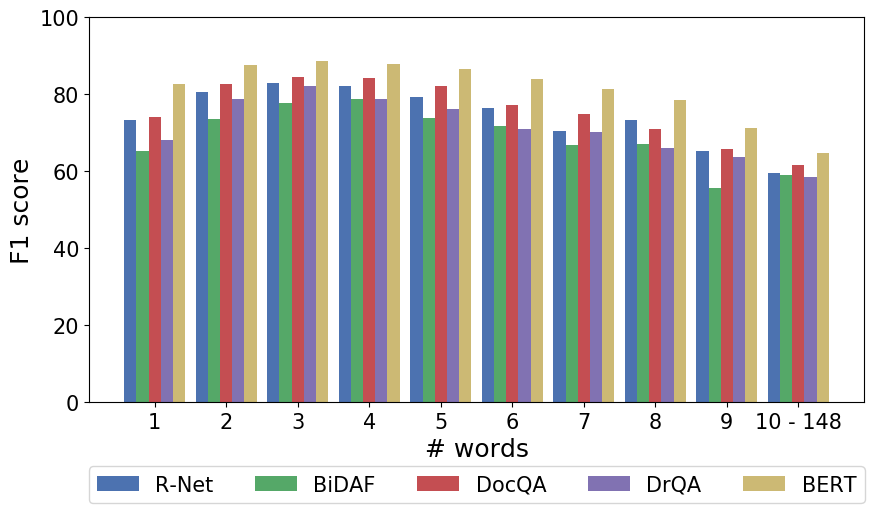}
    \caption{Model performance depending on answer length (\# of words).}
    \label{fig:answer_length_and_f1}
\end{figure}

Following our analysis of the dataset, we break down model scores by the answer types. Tables~\ref{tab:NE_answer_types} and~\ref{tab:NE_whole_answer_types} summarize performance of the models depending on the answers containing named entities of diffirent types. 
Table~\ref{tab:NE_answer_types} represents answers that contain at least one NE,
but which are not necessarily NEs themselves (42.7\% in the test set).
Table \ref{tab:NE_whole_answer_types} represents answers that are NEs (13.8\% in test). 
A common trend for all models is that F1 scores for answers mentioning dates, persons, locations, and organizations are higher than average.
NUMBER is an exception in this regard, probably due to a high variability of contexts might contain numerals both as digits and words. Answers containing \textit{other NEs} also show degraded performance -- probably, again due to their higher diversity and lower counts. The scores are significantly higher when an answer is exactly a NE.
This is in line with previous studies that showed that answers containing NEs are easier to answer, see for example~\cite{systematic-error-analysis}.

For about 48\% of the answers in the testing set that don't contain NEs we were able to derive their syntactic phrase type, see Table~\ref{tab:answers_AOT}. Among them, non-factoid verb phrases stand out as most difficult ones
(all models perform worse on such questions).\footnote{Adverbial phrases appears to be even harder, but they are too few to make reliable conclusions.} 
In contrast, answers expressed as prepositional phrases are easier to answer
compared to both noun and verb phrases.
Noun phrases---most common syntactic units among answers---are second-easiest structure among others to answer.
However, with exception for BERT, F1 scores for noun phrases are lower than average.

The models behave remarkably differently on questions with and without detected misspellings, see Table~\ref{tab:spelling_answer_types}. 
DrQA seems to be most sensible to misspellings: The difference in F1 is almost 8\% (scores are lower for misspelled questions).
DocQA has most stable behavior: The difference in F1 scores is about 2\%. 

Questions with interrogative \lipart-particle represent around 1\% in the whole dataset. 
Although score averages for such small sets are not very reliable, 
the decrease in performance on these questions is quite sharp and consistent for all models: 
It ranges from 8.5\% in F1 points for DocQA to 18.7\% for BiDAF. 
We  hypothesize that these questions are substantially different from other questions and are poorly represented in the training set.

\input{twotables.tex}

\begin{table}[t]
    \centering
    \begin{tabular}{lr}
        \textbf{Category} &  \textbf{\%} \\ 
        \hline
        Incomplete answer & 29\\
        Vague question & 19 \\
        Incorrect answer & 14 \\
        Broad question &  12\\
        Co-reference resolution & 12 \\
        Reasoning & 10 \\
        Misspellings & 6 \\
        No answer & 3 \\
        Yes/no & 3 \\
        Paraphrase & 3 \\
    \end{tabular}
    \caption{Qualitative analysis of 100 difficult questions (questions can be assigned to more than one category).}
    \label{tab:difficult_questions}
\end{table}{}

\begin{table*}[t]
    \centering
    \begin{tabular}{lrrrrrr}
        \textbf{Bigram} & \textbf{\% test} &  \textbf{R-Net} & \textbf{BiDAF} & \textbf{DocQA} & \textbf{DrQA} & \textbf{BERT}\\
         \hline
        \foreignlanguage{russian}{в какой} / in what                                      & 8.62 & 84.2 & 82.7 & 85.8 & 84.6 & 87.7\\
        \foreignlanguage{russian}{как называться} / how is X called    & 2.46 & 84.5 & 74.7 & 81.9 & 78.7 & 89.8\\
        \foreignlanguage{russian}{кто быть} / who was                                     & 1.21 & 81.8 & 71.0 & 83.2 & 78.3 & 89.2\\
        \foreignlanguage{russian}{на какой} / on what                                     & 1.21 & 75.9 & 72.7 & 76.7 & 78.0 & 80.2\\
        \foreignlanguage{russian}{что такой} / what is                & 1.15 & 71.6 & 67.6 & 74.4 & 70.6 & 77.0\\
        \foreignlanguage{russian}{с какой} / with what                                    & 1.01 & 76.6 & 78.3 & 79.4 & 78.4 & 89.9\\
        \foreignlanguage{russian}{для что} / what for                                     & 0.91 & 81.6 & 79.8 & 82.5 & 78.1 & 86.9\\
        \foreignlanguage{russian}{к что} / to what                                        & 0.77 & 90.9 & 82.2 & 86.7 & 88.1 & 90.2\\
        \foreignlanguage{russian}{что являться} / what is             & 0.69 & 84.6 & 88.0 & 93.5 & 87.2 & 93.2\\
        \foreignlanguage{russian}{когда быть} / when was                                  & 0.68 & 79.2 & 82.2 & 84.0 & 86.9 & 92.5\\
        \hline
         Test set               & &     77.8 & 72.2 & 79.5 & 75.0 & 84.8\\
        
    \end{tabular}
    \caption{Model F1 scores depending on questions' leading bigrams (bigrams are lemmatized).}
    \label{tab:bigrams_answer_types}
\end{table*}{}

\begin{table*}[h]
    \centering
    \begin{tabular}{lrrrrrr}
        \textbf{Trigram} & \textbf{\% test} &  \textbf{R-Net} & \textbf{BiDAF} & \textbf{DocQA} & \textbf{DrQA} & \textbf{BERT}\\
         \hline
        \foreignlanguage{russian}{в какой год} / in which year             & 4.39 & 89.4 & 88.8 & 90.4 & 89.9 & 91.0\\
        \foreignlanguage{russian}{в какой город} / in which city           & 0.32 & 87.0 & 88.5 & 87.0 & 83.8 & 92.6\\
        \foreignlanguage{russian}{что представлять себя} / what is         & 0.30 & 58.5 & 46.3 & 51.8 & 52.3 & 58.5\\
        \foreignlanguage{russian}{что происходить с} / what does happen to & 0.28 & 64.6 & 58.6 & 78.2 & 64.1 & 86.8\\
        \foreignlanguage{russian}{с какой год} / starting from which year  & 0.28 & 93.9 & 93.9 & 93.9 & 93.9 & 93.9\\
        \foreignlanguage{russian}{в какой век} / in which century          & 0.26 & 87.7 & 89.2 & 90.8 & 86.9 & 90.8\\
        \foreignlanguage{russian}{в какой период} / in which period        & 0.26 & 86.8 & 83.9 & 88.1 & 82.1 & 86.8\\
        \foreignlanguage{russian}{к что приводить} / what does X lead to   & 0.24 & 83.6 & 72.0 & 75.9 & 79.7 & 70.8\\
        \foreignlanguage{russian}{от что зависеть} / what does X depend on & 0.20 & 78.8 & 73.6 & 79.2 & 84.0 & 92.5\\
        \foreignlanguage{russian}{в какой страна} / in which country       & 0.18 & 97.8 & 97.8 & 91.3 & 94.4 & 100.0\\
        \hline
         Test set               & &     77.8 & 72.2 & 79.5 & 75.0 & 84.8\\
    \end{tabular}    \caption{Model F1 scores depending on questions' leading trigrams (trigrams are lemmatized)}
    \label{tab:trigrams_answer_types}
\end{table*}{}

Due to high variability of starting question n-grams (see Tables~\ref{tab:bigrams_answer_types} and~\ref{tab:trigrams_answer_types}), 
we cannot make reliable statements for all but most frequent ones. 
For these---we can conclude---that model performance is mostly above average. 
There are a few exceptions: 
Notably, some variants of the definition questions \textit{what/who is} are especially hard for BiDAF. 
More concrete \textit{when}-questions appear to be an easier task for all models. 
In the case of trigrams the number of questions of each type is much smaller (recall that the testing set contains around 5,000 questions).
Nevertheless, the scores for most frequent questions \textit{in which year} are much better than the average scores.

Finally, we sampled 100 questions where all models achieved zero F1 score (i.e., they returned a span with no overlap with 
a ground truth answer). 
We manually grouped the sampled questions into the following categories:
\begin{itemize}
\item An entire paragraph or its significant part can be seen as an answer to a \textit{broad/general question}.
\item An answer is \textit{incomplete}, because it contains only a part of an acceptable longer answer. For example for \textit{Q31929} \textit{`Who did notice an enemy airplane?'} only the word \textit{pilots} is marked as ground truth in the context: \textit{On July 15, during a reconnaissance east to Zolotaya Lipa, \underline{pilots} of the 2nd Siberian Corps Air Squadron Lieutenant Pokrovsky and Cornet Plonsky noticed an enemy airplane.}
\item \textit{Vague questions} are related to the corresponding paragraph but seem to be a result of a misinterpretation of the context by a crowdsource worker. For example, in \textit{Q70465} \textit{`What are the disadvantages of TNT comparing to dynamite and other explosives?'} the ground truth answer \textit{`a detonator needs to be used'} is not mentioned as a disadvantage in the paragraph.
A couple of these questions use paronyms of concepts mentioned in the paragraph. For example, \textit{Q46229} asks about \textit{`discrete policy'}, while the paragraph mentions \textit{`discretionary policy'}.
\item  \textit{No answer in the paragraph} and \textit{incorrect answer} constitute more straightforward error cases. 
\item Some questions require \textit{reasoning} and \textit{co-reference resolution}.
\item A small fraction of questions uses \textit{synonyms and paraphrases} that are not directly borrowed from the paragraph. 
\item A relatively large fraction of `difficult' questions contains \textit{misspellings} and imply \textit{yes/no} answers.
\end{itemize} 
The categorization of the sample is summarized in Table~\ref{tab:difficult_questions}. 
One can see from the table that most potential causes of degraded performance can be attributed to poor data quality:
Only 25\% of cases can be explained by a need to deal with linguistic phenomena such as co-reference resolution, reasoning, and paraphrase detection. 

\section{Conclusions}

In this study, we conducted an in-depth analysis of the Russian reading comprehension dataset SberQuAD,
which was created in 2017 
but was neither properly documented nor presented to the scientific community.
SberQuAD creators generally followed a procedure described by the SQuAD authors,
which resulted in similarly high lexical overlap between questions and
sentences with answers.
Our analysis demonstrates that models perform better when such overlap is high.

Despite the similarities between datasets,  all the models perform worse on SberQuAD than on SQuAD, 
which can be attributed to having only a single answer variant and fewer answers that are named entities.
Furthermore, SberQuAD annotations might have been of poorer quality, but it is hard to quantify. 

We believe that the provided analysis 
constitutes an important contribution to research in multilingual QA.
It facilitates further studies
by evaluating off-the-shelf models for reading comprehension task in Russian
and identifying shortcomings related to dataset creation.
The latter  can serve as a guidance for improving/extension of the dataset in the future.

\paragraph{Acknowledgements.} We thank Peter Romov, Vladimir Suvorov, and Ekaterina Artemova (Chernyak) for providing us with details about SberQuAD preparation. We also thank Natasha Murashkina for initial data processing.
    
\section{References}
\bibliographystyle{lrec}
\bibliography{lrec}

\end{document}

%% file: quest_sample.tex
\begin{figure*}[ht]
\begin{center}
\small
    \begin{tabular}{p{16cm}}
\textit{P6418} \begin{otherlanguage*}{russian} Термин Computer science (Компьютерная наука) появился \underline{\textcolor{green}{в 1959 году}} в научном журнале Communications of the ACM, в котором \underline{\textcolor{red}{Луи Фейн}} (Louis Fein) ратовал за создание Graduate School in Computer Sciences (Высшей школы в области информатики) \ldots Усилия Луи Фейна, численного аналитика Джорджа Форсайта и других увенчались успехом: университеты пошли на создание программ, связанных с информатикой, начиная \underline{\textcolor{blue}{с Университета Пердью}} в 1962. \end{otherlanguage*}   \\\hline
\textit{P6418} The term "computer science" appears in a \underline{\textcolor{green}{1959}} article in Communications of the ACM, 
in which \underline{\textcolor{red}{Louis Fein}} argues for the creation of a Graduate School in Computer Science \ldots   Louis Fein's efforts, and those of others such as numerical analyst George Forsythe, were rewarded: universities went on to create such departments, starting with \underline{\textcolor{blue}{Purdue}} in 1962.
 \\\hline 
\textit{\textcolor{green}{Q11870}} \begin{otherlanguage*}{russian}Когда впервые был применен термин Computer science ( Компьютерная наука )? \end{otherlanguage*}\\
\textit{\textcolor{green}{Q11870}} When did the term "computer science" appear? \\\hline
\textit{\textcolor{red}{Q28900}} \begin{otherlanguage*}{russian}Кто впервые использовал этот термин?\end{otherlanguage*} \\
\textit{\textcolor{red}{Q28900}} Who was the first to use this term? \\\hline
\textit{\textcolor{blue}{Q30330}} \begin{otherlanguage*}{russian}Начиная с \underline{каого} учебного заведения стали применяться учебные программы, связанные с информатикой?\end{otherlanguage*}\\
\textit{\textcolor{blue}{Q30330}} Starting with \underline{wich} university were computer science programs created?\\\hline
\end{tabular}
\end{center}
\caption{\label{fig:examples} A sample SberQuAD entry (both the original and the translation): answers are underlined and colored. 
The word \textbf{which} in \textit{Q30330} is misspelled on purpose to reflect the fact that the original has a misspelling.}
\end{figure*}

%% file: datastat_table.tex
\begin{table}[t]
\begin{center}
\begin{tabular}{l@{\hspace{0em}}r@{\hspace{1em}}r}
 &  \begin{tabular}{c}\textit{SberQuAD}\\\textit{train}\\\end{tabular} & 
 \begin{tabular}{c}\textit{SQuAD 1.1}\\\textit{train/dev}\\\end{tabular} \\
\hline
        \# questions & 50,364 &  87,599 / 10,570\\
\# unique paragraphs & 9,080 & 18,896 / 2,067\\\hline
\multicolumn{3}{c}{Number of tokens} \\\hline
avg. paragraph length & 101.7 & 116.6 / 122.8\\
avg. question length & 8.7 & 10.1 / 10.2\\
avg. answer length & 3.7 & 3.16 / 2.9\\
avg. answer position  & 40.5 & 50.9 / 52.9\\\hline
\multicolumn{3}{c}{Number of characters} \\\hline

avg. paragraph length & 753.9 & 735.8 / 774.3\\

avg. question length & 64.4 & 59.6 / 60.0\\

avg. answer length & 25.9 & 20.2 / 18.7\\

avg. answer position & 305.2 & 319.9 / 330.5\\
question-paragraph LCMS & 32.7 & 19.5 / 19.8\\
\end{tabular}
\end{center}
\caption{\label{tab:data_stats} SberQuAD statistics in the \#
of characters and tokens. 
LCMS stands for the longest contiguous matching subsequence.}
\end{table}

%% file: twotables.tex
\begin{table}[t]
    \centering
    \begin{tabular}{l@{\hspace{-0.5em}}r@{\hspace{0.25em}}r@{\hspace{0.25em}}r@{\hspace{0.25em}}r@{\hspace{0.25em}}r@{\hspace{0.25em}}r}
        & \textbf{\% test} &  \textbf{R-Net} & \textbf{BiDAF} & \textbf{DocQA} & \textbf{DrQA} & \textbf{BERT}\\
         \hline
        w/ typos  & 5.7 & 74.1 & 66.7 & 77.5 & 67.5 & 81.1 \\
        correct   & 94.3 & 77.1 & 72.5 & 79.6 & 75.4 & 85.0 \\
        \hline
         Test set               & &     77.8 & 72.2 & 79.5 & 75.0 & 84.8\\
    \end{tabular}
    \caption{Answer quality for misspelled questions.}
    \label{tab:spelling_answer_types}
\end{table}{}

\begin{table}[t]
    \centering
    \begin{tabular}{l@{\hspace{-0.5em}}r@{\hspace{0.25em}}r@{\hspace{0.25em}}r@{\hspace{0.25em}}r@{\hspace{0.25em}}r@{\hspace{0.25em}}r}
        & \textbf{\% test} &  \textbf{R-Net} & \textbf{BiDAF} & \textbf{DocQA} & \textbf{DrQA} & \textbf{BERT}\\
         \hline
        w/ \lipart      & 1.0 & 66.6 & 53.7 & 71 & 57.5 & 73.3 \\
        other   & 99.0 & 77.9 & 72.4 & 79.6 & 75.2 & 84.9 \\
        \hline
        Test set        & &     77.8 & 72.2 & 79.5 & 75.0 & 84.8\\
    \end{tabular}
    \caption{Yes/no (\lipart-particle) questions.}
    \label{tab:yes_no_question}
\end{table}{}